\documentclass{article}

\usepackage{microtype}
\usepackage{graphicx}
\usepackage{subfigure}
\usepackage{booktabs} 
\usepackage{array}
\usepackage{amsmath}

\usepackage{hyperref}

\usepackage[accepted]{icml2019}

\icmltitlerunning{Spatially-Coupled Neural Network Architectures}

\begin{document}

\twocolumn[
\icmltitle{Spatially-Coupled Neural Network Architectures}



\icmlsetsymbol{equal}{*}

\begin{icmlauthorlist}
\icmlauthor{Arman Hasanzadeh}{equal,to}
\icmlauthor{Nagaraj T. Janakiraman}{equal,to}
\icmlauthor{Vamsi K. Amalladinne}{to}
\icmlauthor{Krishna R. Narayanan}{to}
\end{icmlauthorlist}

\icmlaffiliation{to}{Department of Electrical and Computer Engineering, Texas A\&M University, College Station, Texas, USA}

\icmlcorrespondingauthor{Arman Hasanzadeh}{armanihm@tamu.edu}

\icmlkeywords{Machine Learning, Deep Learning, Sparse Coding}

\vskip 0.3in
]



\printAffiliationsAndNotice{\icmlEqualContribution} 

\begin{abstract}
In this work, we leverage advances in sparse coding techniques to reduce the number of trainable parameters in a fully connected neural network.
While most of the works in literature impose $\ell_1$ regularization, DropOut or DropConnect techniques to induce sparsity, our scheme considers feature importance as a criterion to allocate the trainable parameters (resources) efficiently in the network.
Even though sparsity is ensured, $\ell_1$ regularization requires training on all the resources in a deep neural network.
The DropOut/DropConnect techniques reduce the number of trainable parameters in the training stage by dropping a random collection of neurons/edges in the hidden layers.
However, both these techniques do not pay heed to the underlying structure in the data when dropping the neurons/edges.
Moreover, these frameworks require a storage space equivalent to the number of parameters in a fully connected neural network.
We address the above issues with a more structured architecture inspired from spatially-coupled sparse constructions.
The proposed architecture is shown to have a performance akin to a conventional fully connected neural network with dropouts, and yet achieving a $94\%$ reduction in the training parameters.
Extensive simulations are presented and the performance of the proposed scheme is compared against traditional neural network architectures.
\end{abstract}

\section{Introduction}
Deep neural networks excel at many tasks but usually suffer form two problems: 1) they are cumbersome, and difficult to deploy on embedded devices \cite{crowley2018pruning} and 2) they tend to overfit the training data \cite{hawkins2004problem}. Several approaches have been proposed to overcome each of them. DropOut \cite{srivastava2014dropout}, DropConnect \cite{wan2013regularization} and wight-decay \cite{krogh1992simple} are a few examples of techniques that try to overcome overfitting. DropOut drops a random subset of neurons and all the edges connected to them during the training phase while in DropConnect a random subset of edges are dropped during the training phase. Weight-decay method uses $\ell_2$ regularization to enforce uniformity among weights. However, all of these approaches induce sparsity only during the training phase and testing is performed on the averaged network which is not sparse.
Hence, there is no reduction in the memory usage.

Pruning methods try to compress a deep neural network to be more memory efficient. A simple, yet popular technique which uses hard thresholding to prune the weights close to zero has received significant attention \cite{reed1993pruning}. Weight-elimination \cite{weigend1991generalization}, group LASSO regularizer \cite{scardapane2017group} and $L_{1/2}$ regularizer \cite{li2018smooth} are examples of pruning techniques. Even thought all these approaches have shown a lot of promise in sparsifying the network, they are computationally taxing schemes, since all the weights of the network need to be optimized over training data.

In this work, first, we empirically show that $\ell_1$ regularization results in a paradigm where the out degree of a neuron is representative of its importance. Then, we propose a random sparse neural network which trains on a far fewer parameters than a fully connected deep neural network without much degradation in the test performance. Our approach uses feature importance to order the inputs in the first layer and inputs with lesser importance are allotted lesser trainable parameters. This sparsity structure is maintained across all the layers by using spatially-coupled sparse constructions (inspired by spatially-coupled LDPC codes) to maintain block sparsity. Our proposed architecture is pruned before training and avoids overfitting. Hence, it is both computationally efficient and memory efficient.

\section{Proposed Sparse Construction}
In most of the learning problems at hand, features are transformed into a space that separates the data well to aid the learning task. But traditional neural networks don't take advantage of the inherent ordering in the features which is based on an importance measure. Our proposed architecture is a feed-forward neural network that is designed to take advantage of the side information in the input features to allocate the trainable parameters efficiently. 
The first step in our model is to transform and rank input features based on their importance. Two of such transformations that well suit our model are principal component analysis (PCA) \cite{wold1987principal} and random forest (RF) feature importance \cite{liaw2002classification}. 
After applying one of the feature transformation methods to the input features, we get transformed input $\mathbf{x} = [x_1, x_2, \cdots, x_N ]$
where the elements in $\mathbf{x}$ are ordered in the decreasing order of importance measure specific to the transformation. In the next section first we define spatially-coupled (SC) layer and then show how to use feature importnace information in a SC layer.

\subsection{Spatially-Coupled (SC) Layer}
First, we define random sparse (RSP) layer which is the building block of a SC layer.
Let $\mathbf{z}^{(l)}$ denote the output vector of the $l$-th layer of the neural network. Also assume that $\mathbf{W}^{(l)}$ and $\mathbf{b}^{(l)}$ denote the weights and biases at layer $l$, respectively. The output of RSP layer $l$ is given by

\begin{equation}
\label{eqn:layer_out}
    \mathbf{z}^{(l)} = \text{act}(\mathbf{W}^{(l)} \mathbf{z}^{(l-1)} + \mathbf{b}^{(l)})\\
\end{equation}

where $\text{act}(.)$ is the activation function. 
Although the mathematical formulation in (\ref{eqn:layer_out}) is similar to traditional fully connected (FC) layer, the fundamental difference is that $\text{supp}(\mathbf{W}^{(l)})$ is the binary adjacency matrix of a random bipartite graph (Tanner graph) as opposed to all ones matrix in the FC case. 
More specifically, RSP construction imposes sparsity in the bipartite graph between layers, in which each edge denotes a weight between two neurons in layers $(l-1)$ and $l$, in the same way as how low density parity check (LDPC) codes are constructed \cite{gallager1962low}.

Assume that $d_{i}^{(l)}$ shows the out degree of $i$-th neuron at layer $l$, $[K]$ denotes the set of integers $\{1, \cdots, K\}$ and $N^{(l)}$ represents the number of neurons at layer $l$. Then, the RSP layer is constructed as follows.
\begin{enumerate}\itemsep 0.5pt
    \item Pick a degree distribution $\lambda(d;\mathcal{P})$ parameterized by the set of parameters $\mathcal{P}$.
    \item Draw $N^{(l-1)}$ i.i.d samples for degrees $\{d_i^{(l-1)}\}_{i=1}^{N^{(l-1)}}$ from the chosen degree distribution $\lambda(d;\mathcal{P})$.
    \item For each neuron $i$ in layer $(l-1)$, uniformly pick $d_i^{(l-1)}$ unique neurons denoted by $J \subseteq [N^{(l)}]$ from layer $l$ and connect $i$ to all neurons in $J$.
\end{enumerate}

Notice that unlike FC layer, not every neuron in layer $l-1$ is connected to every neuron in layer $l$. In other words, the out degree of $i$-th neuron at layer $l-1$ is much less than the number of nodes in layer $l$, i.e. $d_i^{(l-1)}\ll N^{(l)}$.

\begin{figure}[t]
\centering
\includegraphics[height=0.75\linewidth]{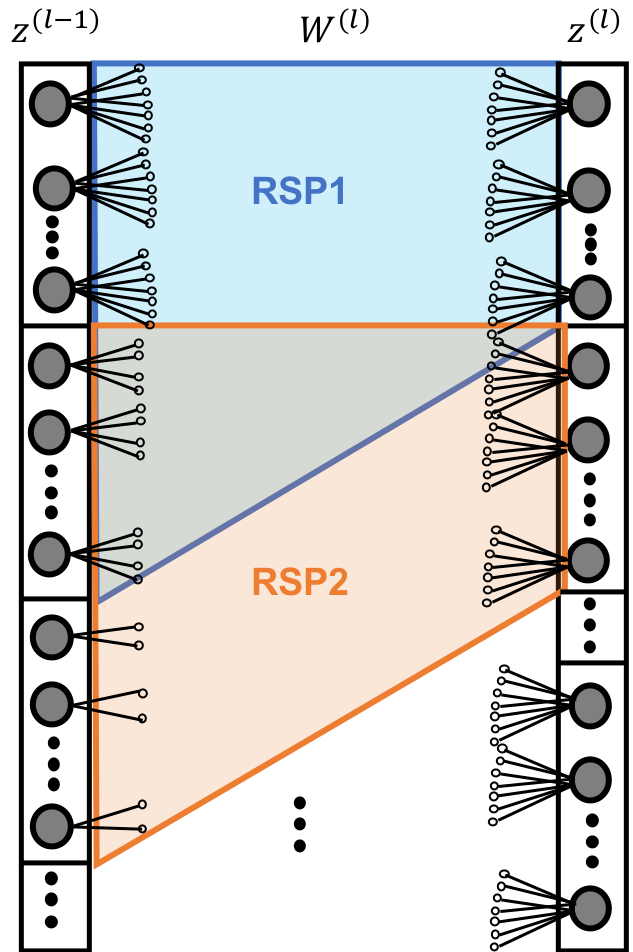}
\caption{Schematic of a spatially-coupled layer with left regular random blocks and $r^{(l)}=2$.}
\label{fig:sc_lay}
\end{figure}

Given the construction of a RSP layer, now we define the SC layer. Here, the neurons in each layer $l$ are partitioned into $B^{(l)}$ blocks of equal size and the neurons in a block of layer $l$ are randomly connected (locally) to neurons in a few adjacent blocks of layer $l-1$. 
The number of adjacent blocks that participate in the local connections at layer $l$ is called the receptive field of the layer and denoted by $r^{(l)}$ (see Figure \ref{fig:sc_lay}).
More specifically, the construction of SC layer is done as follows:
\begin{enumerate} \itemsep 0.5pt
    \item Consider a window of $r^{(l-1)}$ adjacent blocks with block indices $\{i, \cdots , i+r^{(l-1)}-1\}$ from layer $l-1$ and block index $i$ from layer $l$ and construct a RSP layer locally. 
    \item Repeat step 1 for each of the blocks and choose a random instance of RSP for each of those windows.
\end{enumerate}

We choose a simple left regular degree distribution within each block. As pointed out before, we allocate resources (trainable parameters) according to the importance in features. The degree of each neuron is equivalent to the amount of resources that is allocated to that neuron. As our input is ordered based on some importance measure, we allocate high degree to the blocks with higher important features and low degree to the blocks with lower important features. 
By construction, the intermediate layers also have the same ordering of feature importance and hence at each layer, we allocate degrees proportional to the importance measure.

\begin{figure}[t]
\centering
\includegraphics[width=\columnwidth]{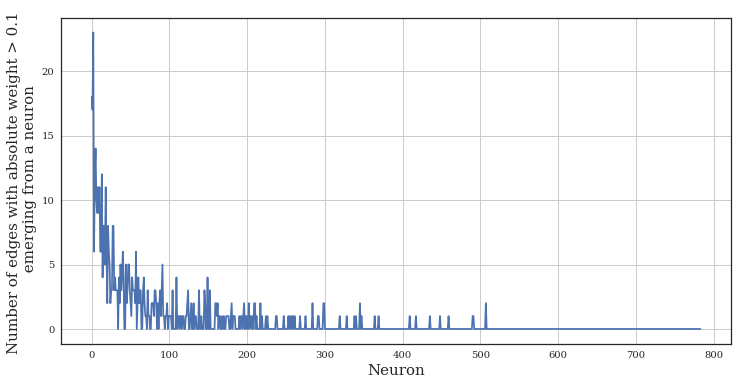}
\label{fig:sub3}
\caption{This plot illustrates the combination of PCA and $\ell_1$ regularization techniques for the input layer. The number of edges  with absolute weight greater than 0.1 emerging from a neuron decays as the feature importance decays when $\ell_1$ regularization is used.}
\label{fig:l1_num}
\end{figure}

\section{Experiments and Discussion}
We trained spatially-coupled neural network for classification problem on fashion MNIST dataset \cite{fashionmnist}. 
Fashion MNIST data set consist of 70000 samples of 28$\times$28 gray-scale images, each one associated with a label from 10 classes. 
We vectorized the input samples to a vectors of length 784, transformed (in case of PCA) and reordered the features based on decreasing order of importance.

In all of our experiments, we deployed neural networks with 5 hidden layers with 784 neurons each and an output layer with 10 neurons. We used sigmoid activation function, cross-entropy loss with $\ell_2$ regularization and regularization parameter of $5\times10^{-5}$ except for FC architecture.
In FC architecture, we used DropOut with keeping probability of 0.5. We also compared our method with RSP construction with left regular degree distribution. The degree of each node is set to be 53. We form our proposed SC neural network with 8 blocks in all layers. The RSP construction of each block is left regular. 
The out degrees of the neurons in blocks are set to be \{98, 130, 98, 49, 20, 10, 10, 10\}. We note that the number of parameters in RSP and SC constructions are approximately 93\% less than the fully connected case.

The findings reported in Figure \ref{fig:l1_num} fortify the proposed framework of allocating more resources to features of higher importance.
We trained a FC neural network classifier (with the same architecture as mentioned above) with $\ell_1$ regularization of $10^{-4}$ on input layer weights and $10^{-5}$ on other weights. PCA was used to sort features in descending order.
The number of edges with absolute weight greater than 0.1 (a representative of contributing edges) emerging from an input neurons is plotted in Figure \ref{fig:l1_num}.
It can be seen that it decays rapidly as the importance of features goes down which validates our choice of SC graph constructions. 

\begin{table}[t]
\centering
\caption{Accuracy of various neural network construction methods for fashion MNIST classification.}
\vspace{0.1in}
\resizebox{\columnwidth}{!}{
\scalebox{1} {
\normalsize{
\begin{tabular}{c|c|c|c}
\toprule
Feature Importance & Input Ordering & NN Construction & Accuracy\\ \hline
\midrule
& & SC  & \textbf{87.18}\% \\
 \cline{3-4}
& Descending & RSP & 84.33\% \\ 
 \cline{3-4}
& & FC & 86.78\% \\ 
 \cline{2-4}
PCA
 & & SC  & 10.00\% \\ 
 \cline{3-4}
& Ascending & RSP & 84.33\% \\ 
 \cline{3-4}
& & FC & \textbf{86.78}\% \\ 
\hline
\hline
 & & SC  & \textbf{86.40}\% \\ 
 \cline{3-4}
& Descending & RSP & 84.54\% \\ 
 \cline{3-4}
& & FC & 86.06\% \\ 
\cline{2-4}
RF
 & & SC  & 85.26\% \\ 
 \cline{3-4}
& Ascending & RSP & 84.54\% \\ 
 \cline{3-4}
& & FC & \textbf{86.06}\% \\ 

\bottomrule
\end{tabular}
}
}\label{tbl: res}
}
\end{table}

Table \ref{tbl: res} summarizes the results of classification task. If the input to the models is ordered PCA features, spatially-coupled neural network shows the best accuracy, 87.18\%. Comparing SC with FC neural network shows that FC, even with $\ell_2$ regularization and DropOut, tend to overfit the data because of the huge number of parameters in the model. On the other hand, adding sparsity to the model naively like RSP with left regular degree neural network degrades the performance. By allocating the trainable parameters efficiently and cleverly, we can reach high accuracy with very sparse networks. The same pattern exist for random forest feature importance (RF) too, in which SC outperforms the other two methods.

To show that ordering of features is crucial in SC construction, we repeated the experiments with reversed order of features, i.e. assigning high degrees to less important features and vice versa. It can be seen that in reverse PCA case, SC shows a very poor performance while the other methods maintain their performance as they are permutation invariant. In RF reverse case, SC is very close to the best performance. The difference in the two cases, that causes drastic change in performance, is the quantization of feature importance. The PCA has a small number of very important features (few high variance features) and large number of feature which are not important (many low variance features), thus assigning a very low degree to all of the high variance features degrading the accuracy substantially. However, RF reorders the input which tend to give us many equally important feature and some less important features. Therefore, all of the important features are not diminished and some of them will have high degree in SC model.

One important property of SC construction is that it is preserves the feature importance ordering throughout the network. We validated this empirically by measuring the feature importance at each layer. Besides better interpretability of the model compared to FC neural networks, a nice application of this property is that at every layer we can prune the lower blocks after training while maintaining the overall accuracy. This can lead us to a highly sparse structure which can reduce the model complexity even more than 95\% with approximately the same performance. An avenue for future work is how to learn these class of transformation that respects the network using fully connected layers.

\bibliography{References.bib}

\begin{thebibliography}{13}
\providecommand{\natexlab}[1]{#1}
\providecommand{\url}[1]{\texttt{#1}}
\expandafter\ifx\csname urlstyle\endcsname\relax
  \providecommand{\doi}[1]{doi: #1}\else
  \providecommand{\doi}{doi: \begingroup \urlstyle{rm}\Url}\fi

\bibitem[Crowley et~al.(2018)Crowley, Turner, Storkey, and
  O'Boyle]{crowley2018pruning}
Crowley, E.~J., Turner, J., Storkey, A., and O'Boyle, M.
\newblock Pruning neural networks: is it time to nip it in the bud?
\newblock \emph{arXiv preprint arXiv:1810.04622}, 2018.

\bibitem[Gallager(1962)]{gallager1962low}
Gallager, R.
\newblock Low-density parity-check codes.
\newblock \emph{IRE Transactions on information theory}, 8\penalty0
  (1):\penalty0 21--28, 1962.

\bibitem[Hawkins(2004)]{hawkins2004problem}
Hawkins, D.~M.
\newblock The problem of overfitting.
\newblock \emph{Journal of chemical information and computer sciences},
  44\penalty0 (1):\penalty0 1--12, 2004.

\bibitem[Krogh \& Hertz(1992)Krogh and Hertz]{krogh1992simple}
Krogh, A. and Hertz, J.~A.
\newblock A simple weight decay can improve generalization.
\newblock In \emph{Advances in neural information processing systems}, pp.\
  950--957, 1992.

\bibitem[Li et~al.(2018)Li, Zurada, and Wu]{li2018smooth}
Li, F., Zurada, J.~M., and Wu, W.
\newblock Smooth group l1/2 regularization for input layer of feedforward
  neural networks.
\newblock \emph{Neurocomputing}, 314:\penalty0 109--119, 2018.

\bibitem[Liaw et~al.(2002)Liaw, Wiener, et~al.]{liaw2002classification}
Liaw, A., Wiener, M., et~al.
\newblock Classification and regression by randomforest.
\newblock \emph{R news}, 2\penalty0 (3):\penalty0 18--22, 2002.

\bibitem[Reed(1993)]{reed1993pruning}
Reed, R.
\newblock Pruning algorithms-a survey.
\newblock \emph{IEEE transactions on Neural Networks}, 4\penalty0 (5):\penalty0
  740--747, 1993.

\bibitem[Scardapane et~al.(2017)Scardapane, Comminiello, Hussain, and
  Uncini]{scardapane2017group}
Scardapane, S., Comminiello, D., Hussain, A., and Uncini, A.
\newblock Group sparse regularization for deep neural networks.
\newblock \emph{Neurocomputing}, 241:\penalty0 81--89, 2017.

\bibitem[Srivastava et~al.(2014)Srivastava, Hinton, Krizhevsky, Sutskever, and
  Salakhutdinov]{srivastava2014dropout}
Srivastava, N., Hinton, G., Krizhevsky, A., Sutskever, I., and Salakhutdinov,
  R.
\newblock Dropout: a simple way to prevent neural networks from overfitting.
\newblock \emph{The Journal of Machine Learning Research}, 15\penalty0
  (1):\penalty0 1929--1958, 2014.

\bibitem[Wan et~al.(2013)Wan, Zeiler, Zhang, Le~Cun, and
  Fergus]{wan2013regularization}
Wan, L., Zeiler, M., Zhang, S., Le~Cun, Y., and Fergus, R.
\newblock Regularization of neural networks using dropconnect.
\newblock In \emph{International Conference on Machine Learning}, pp.\
  1058--1066, 2013.

\bibitem[Weigend et~al.(1991)Weigend, Rumelhart, and
  Huberman]{weigend1991generalization}
Weigend, A.~S., Rumelhart, D.~E., and Huberman, B.~A.
\newblock Generalization by weight-elimination with application to forecasting.
\newblock In \emph{Advances in neural information processing systems}, pp.\
  875--882, 1991.

\bibitem[Wold et~al.(1987)Wold, Esbensen, and Geladi]{wold1987principal}
Wold, S., Esbensen, K., and Geladi, P.
\newblock Principal component analysis.
\newblock \emph{Chemometrics and intelligent laboratory systems}, 2\penalty0
  (1-3):\penalty0 37--52, 1987.

\bibitem[{Xiao} et~al.(2017){Xiao}, {Rasul}, and {Vollgraf}]{fashionmnist}
{Xiao}, H., {Rasul}, K., and {Vollgraf}, R.
\newblock {Fashion-MNIST: a Novel Image Dataset for Benchmarking Machine
  Learning Algorithms}.
\newblock \emph{arXiv e-prints}, art. arXiv:1708.07747, Aug 2017.

\end{thebibliography}
\bibliographystyle{icml2019}
\end{document}